\documentclass[runningheads]{llncs}

\usepackage{accv}

\usepackage{accvabbrv}

\usepackage{graphicx}
\usepackage{booktabs}
\usepackage{float}
\usepackage[accsupp]{axessibility}  
\usepackage{algorithm}
\usepackage{algpseudocode}

\usepackage{hyperref}

\usepackage{orcidlink}

\begin{document}

\title{DLC: Dynamic Loss Controller for Multi-Objective Image Restoration}

\titlerunning{DLC for Multi-Objective Image Restoration}

\author{
  Jaewan Ko\orcidlink{0009-0004-4866-7258} \and
  Janghoon Choi\thanks{Corresponding Author}\orcidlink{0000-0002-0932-6213}
}

\authorrunning{J. Ko and J. Choi}

\institute{
  Major in Data Science Convergence, Graduate School of Data Science\\
  Kyungpook National University, Daegu, Republic of Korea\\
  \email{\{asdf8800, jhchoi09\}@knu.ac.kr}
}

\maketitle

\begin{abstract}
  In this paper, we introduce a metric-guided dynamic loss controller (DLC) for multi-objective image restoration. Conventional image restoration pipelines usually train with a fixed weighted combination of multiple losses, without changing the relative importance of fidelity, perceptual similarity, and no-reference quality during optimization.
  DLC is an architecture- and loss-term-agnostic training-time controller: it does not modify the restoration architecture or introduce new differentiable loss terms, but dynamically reweights the existing training losses. During training, DLC periodically evaluates the current model on a small fixed feedback subset and uses the resulting quality metrics to update the loss-weight vector through an LLM-based controller.
  Because DLC operates on existing loss terms rather than task-specific architectures, the same controller formulation can be instantiated across diverse image restoration training pipelines.
  We evaluate DLC on three restoration domains: low-light image enhancement, deraining, and real-world super-resolution, using both reference-based and no-reference quality metrics. Across these settings,
  DLC considers metric-dependent trade-offs during optimization and guides training toward balanced operating points across fidelity and perceptual quality.
  The results show that DLC can move models toward more favorable operating points across different restoration domains, supporting its role as a practical plug-in controller for multi-objective image restoration.
\end{abstract}

\section{Introduction}
\label{sec:intro}

Deep image restoration models are commonly optimized with a weighted combination of multiple loss terms~\cite{johnson2016perceptual,ledig2017photo,wang2018esrgan,wei2018deep,zamir2020learning,zamir2021multi,zamir2022restormer,chen2022simple}. Various loss functions have been used: pixel-level losses encourage fidelity to a reference image, structural losses preserve local layout and contrast, perceptual losses promote realistic output through feature-level similarity, and task-specific objectives drive the behavior required by a particular degradation model. In most training pipelines, however, the relative weights of these losses are selected before training and then kept fixed until convergence. This design is convenient and reproducible, but it assumes that one static weighting scheme remains appropriate throughout the entire optimization process.

This assumption is restrictive because restoration quality is inherently multi-objective. Fidelity-oriented metrics such as PSNR and SSIM~\cite{wang2004image} reward accurate reconstruction of the reference signal, while perceptual and no-reference metrics may favor sharper textures, stronger contrast, or more natural image statistics~\cite{zhang2018unreasonable,ding2020image,mittal2012making,mittal2012no}. Improving one objective can therefore weaken another: conservative outputs often obtain high PSNR but appear over-smoothed, whereas stronger perceptual emphasis may recover fine details while introducing artifacts or reducing pixel-level fidelity~\cite{blau2018perception}. This perception--distortion tension has been widely discussed in image restoration and motivates the need to consider balanced operating points, which here refer to favorable fidelity--perception trade-offs rather than formally defined Pareto optima~\cite{blau2018perception}.

\begin{figure}[t]
  \centering
  \includegraphics[width=0.92\linewidth,trim=0cm 1cm 0cm 2.5cm,clip]{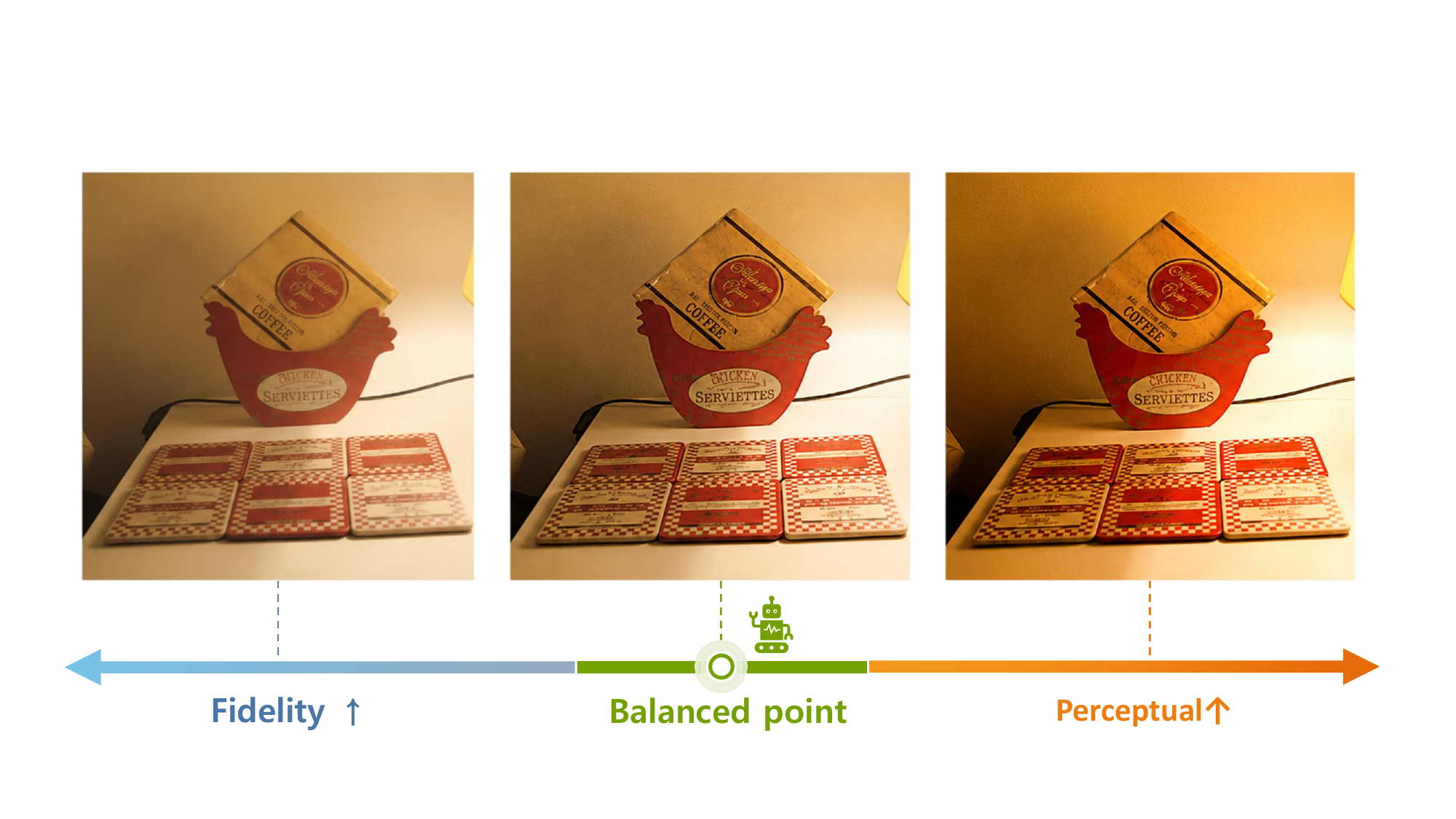}
  \caption{
    Conceptual illustration of the fidelity--perceptual trade-off in image restoration. Fidelity-oriented optimization tends to preserve reference consistency, while perceptual-oriented optimization emphasizes visual sharpness and detail. DLC aims to guide training toward a balanced operating point through metric-guided loss reweighting.
  }

  \label{fig:tradeoff_concept}
\end{figure}

Several adaptive loss-weighting methods have been proposed for multi-task or multi-objective learning, often using task uncertainty, training-loss changes, gradient magnitudes, validation feedback, learned schedules, or multi-objective gradient information as the adjustment signal~\cite{kendall2018multi,chen2018gradnorm,liu2019end,ren2018learning,xu2018autolosslearningdiscreteschedules,sener2018multi,yu2020gradient}. These signals are useful, but they do not directly represent the evaluation criteria used in restoration benchmarks. Restoration papers are usually judged by a heterogeneous metric set that may include reference fidelity, reference-based perceptual similarity, no-reference quality, and distributional realism. This creates a practical gap: the quantities used to tune the training objective are often different from the quantities used to evaluate the final model.

\begin{figure}[t]
  \centering
  \includegraphics[width=\linewidth, trim=0 1.2cm 0 2.0cm, clip]{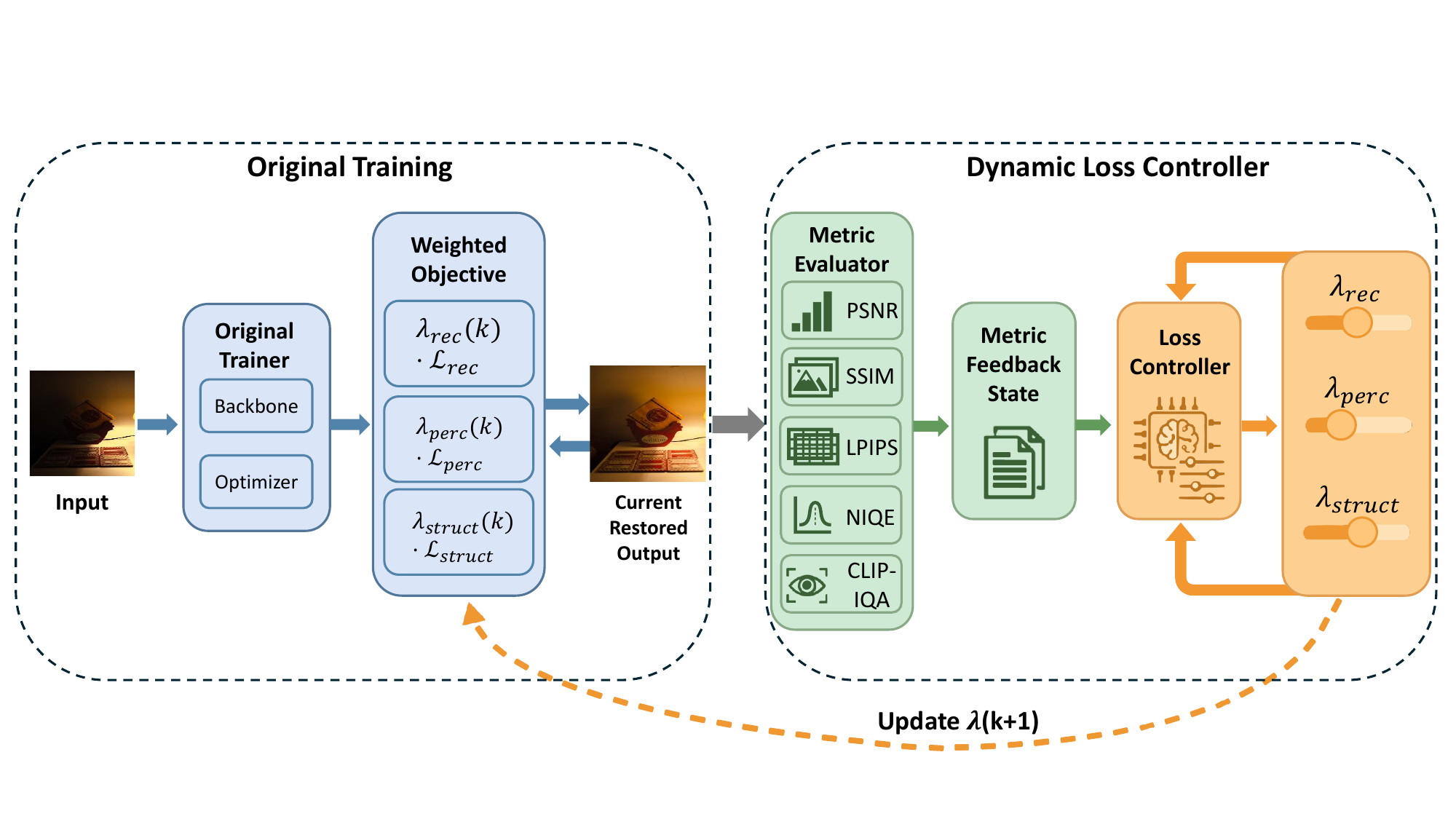}
  \caption{
    Overview of DLC. At intervention step $k$, the current restored output is evaluated by task-relevant metrics to construct a feedback state. The loss controller then produces the next loss-weight vector $\boldsymbol{\lambda}_{k+1}$ for the following training interval.
  }
  \label{fig:dlc_overview}

\end{figure}

Herein, we propose DLC, a metric-guided dynamic loss controller for multi-objective training. Rather than designing a new backbone or introducing a new differentiable metric loss, DLC treats loss weighting itself as a feedback-driven training decision. During training, DLC periodically evaluates the current model on a fixed feedback subset sampled from the training distribution. It then constructs a structured feedback state from the current metric profile, recent metric changes, and loss-weight history. An LLM-based controller maps this state to the next loss-weight vector for the following training interval~\cite{yang2024large,liu2024large,hili2025llms}. In this way, DLC converts heterogeneous quality-metric feedback into adaptive loss reweighting while preserving the original model, optimizer, data pipeline, and set of training losses. As illustrated in Fig.~\ref{fig:dlc_overview}, DLC adds a metric-guided loss-control branch to the original restoration training loop.

To validate that our proposed DLC operates over the existing loss space of each restoration model, we evaluate DLC on three restoration domains. In low-light enhancement~\cite{zhang2025cwnet}, it adjusts pixel, perceptual, and SSIM loss weights. In deraining~\cite{dong2025channel}, it adjusts reconstruction, consistency, perceptual, and adversarial-related loss multipliers. In real-world super-resolution~\cite{sun2025pixel}, it adjusts L2, LPIPS, and CSD objective weights. These instantiations differ in architecture, supervision, and loss design, but share the same feedback-to-action formulation. This makes DLC a training-time plug-in rather than a task-specific architectural modification.

For quantitative evaluation, we evaluate DLC on three restoration domains and seven benchmark settings. In low-light enhancement, DLC improves multiple fidelity, perceptual, and no-reference metrics depending on the dataset. In deraining, DLC shifts the balance between fidelity-oriented and perceptual quality measures. In real-world super-resolution, DLC improves reference fidelity and reference-based perceptual metrics on DRealSR and RealSR. These results do not show that one method improves every metric simultaneously. Instead, they show that metric-guided loss control can steer training toward useful operating points while exposing the remaining trade-offs among metrics.

\noindent\textbf{Contributions.}
\begin{itemize}
  \item We formulate dynamic loss reweighting as a metric-guided training decision for multi-objective restoration, connecting training-time loss control to quality-metric feedback measured on a fixed feedback subset.
  \item We introduce DLC, an LLM-driven plug-in controller that converts metric feedback and recent training history into adaptive loss-weight updates.
  \item We instantiate DLC in three heterogeneous pipelines: low-light enhancement, deraining, and real-world super-resolution.
  \item We provide quantitative and trade-off-oriented evidence across seven benchmarks, showing that DLC can improve key fidelity or perceptual metrics without modifying the underlying backbone architectures.
\end{itemize}

\section{Related Work}
\label{sec:related_work}

\subsection{Composite Losses in Image Restoration}

Modern image restoration models are commonly trained with a weighted combination of multiple loss terms. Pixel-wise losses encourage signal fidelity, structural losses preserve local layout and contrast, perceptual losses promote feature-level similarity, and adversarial or task-specific objectives are often introduced to improve realism or enforce degradation-specific constraints~\cite{johnson2016perceptual,ledig2017photo,wang2018esrgan,wei2018deep}. This composite-loss design is widely used across restoration tasks, including low-light image enhancement, deraining, and real-world super-resolution~\cite{zhang2019kindling,guo2020zero,wang2022low,dong2025channel,wang2021real,sun2025pixel,zhang2025cwnet}. However, the relative coefficients of these losses are typically selected before training and then kept fixed throughout optimization. Such a fixed configuration implicitly assumes that the same trade-off among fidelity, perceptual quality, and task-specific behavior remains appropriate during the entire training process. DLC addresses this fixed-weight assumption by using the same set of training losses as the baseline while dynamically adjusting their weights according to metric feedback.

\subsection{Dynamic Loss Weighting for Multi-Objective Training}

Dynamic loss weighting has been studied in multi-task and multi-objective learning~\cite{kendall2018multi,chen2018gradnorm,liu2019end,ren2018learning,xu2018autolosslearningdiscreteschedules,sener2018multi,yu2020gradient}. Existing approaches adjust loss weights using signals such as task uncertainty~\cite{kendall2018multi}, changes in training loss~\cite{liu2019end}, gradient magnitudes~\cite{chen2018gradnorm}, or multi-objective gradient information~\cite{sener2018multi,yu2020gradient}. These methods show that static loss coefficients can be suboptimal when multiple objectives are optimized jointly. However, most existing weighting strategies rely on internal training signals, such as losses or gradients, rather than on the heterogeneous quality metrics used to evaluate image restoration models. In restoration, the final model is often judged by a mixture of fidelity, reference-based perceptual, no-reference, and distributional metrics. DLC differs from conventional loss-balancing methods by treating these quality metrics as training-time feedback for loss-weight control.

\subsection{Quality Metrics for Image Restoration Feedback}

Image restoration methods are evaluated with diverse metrics because no single score captures all aspects of visual quality. PSNR and SSIM~\cite{wang2004image} are commonly used to measure pixel-level fidelity and structural agreement. Reference-based perceptual metrics such as LPIPS~\cite{zhang2018unreasonable} and DISTS~\cite{ding2020image} estimate perceptual similarity in learned or feature-based spaces. No-reference metrics such as NIQE~\cite{mittal2012making}, BRISQUE~\cite{mittal2012no}, CLIP-IQA~\cite{wang2023exploring}, MUSIQ~\cite{ke2021musiq}, and MANIQA~\cite{yang2022maniqa} estimate image quality without a ground-truth reference, while distributional metrics such as FID~\cite{heusel2017gans} are often used to assess realism. These metrics do not always agree with one another. Improving distortion-oriented fidelity can lead to over-smoothed outputs, whereas emphasizing perceptual quality may introduce artifacts or reduce reference fidelity. DLC is motivated by this metric diversity. Instead of collapsing all metrics into a single scalar reward, DLC preserves the multi-metric profile and uses it as structured feedback for adaptive loss reweighting.

\subsection{LLM-Based Optimization and Loss Control}

Large language models~(LLMs) have recently been explored as optimization and decision-making modules beyond text generation. Optimization by PROmpting (OPRO) formulates optimization as a prompting process, where an LLM proposes new candidate solutions from previous solutions and their scores~\cite{yang2024large}. AgentHPO uses an LLM agent to automate hyperparameter optimization by processing task information and iteratively refining configurations based on historical trials~\cite{liu2024large}. Recent work also investigates whether LLMs can act as in-context meta-learners for model and hyperparameter selection from dataset metadata~\cite{hili2025llms}. TextGrad further explores text-based feedback as an optimization signal for improving components of compound AI systems~\cite{yuksekgonul2024textgrad}.

DLC uses an LLM in a narrow and operational role. The LLM does not generate restored images, synthesize labels, or design a new architecture. Instead, it receives a structured feedback state consisting of metric values, metric directions, recent metric changes, current loss weights, and adjustable loss terms. The controller then returns the next loss-weight configuration within the existing set of training losses. This framing makes the LLM a training-time loss controller rather than a generative component. The main contribution of DLC is therefore a metric-guided feedback loop that connects restoration quality evaluation to adaptive loss reweighting.

\section{Method}
\label{sec:method}

\subsection{Problem Formulation}
\label{sec:problem_formulation}

Let $f_{\theta}$ denote a restoration model with parameters $\theta$. Most restoration pipelines train the model with a weighted combination of multiple loss terms,
\begin{equation}
  \mathcal{L}(\theta; \boldsymbol{\lambda})
  =
  \sum_{i=1}^{N} \lambda_i \mathcal{L}_i(\theta),
\end{equation}
where $\mathcal{L}_i$ is an existing loss term in the original training pipeline and $\lambda_i$ is its corresponding weight. The loss terms may represent different objectives, such as pixel fidelity, structural consistency, perceptual similarity, adversarial realism, or task-specific constraints. Standard training uses a fixed weight vector $\boldsymbol{\lambda}$ throughout optimization. This means that the same relative importance among all losses is imposed from the beginning to the end of training.

DLC changes only this fixed-weight assumption. We divide training into a sequence of intervals. During interval $k$, the model is optimized with the current loss-weight vector $\boldsymbol{\lambda}_k$. Within this interval, the original optimizer updates the model parameters using the weighted objective:
\begin{equation}
  \theta_{t+1}
  =
  \theta_t
  -
  \eta_t
  \nabla_{\theta}
  \mathcal{L}(\theta_t; \boldsymbol{\lambda}_k),
  \qquad
  t \in [kT, (k+1)T),
\end{equation}
where $T$ denotes the intervention interval and $\eta_t$ is the learning rate at training step $t$. Thus, $\boldsymbol{\lambda}_k$ is kept fixed within each interval and updated only at discrete intervention steps.

After the interval, the current model is evaluated on a fixed feedback subset, producing a set of feedback metrics. For each metric $M_j$, DLC records both its value $m_k^{(j)}$ and its optimization direction $d_j \in \{\uparrow,\downarrow\}$, where $\uparrow$ denotes a higher-is-better metric and $\downarrow$ denotes a lower-is-better metric. A controller then maps the metric state and recent training history to the next loss-weight vector:
\begin{equation}
  \mathbf{s}_k =
  \left(
    \{(m_k^{(j)}, d_j)\}_{j=1}^{J},
    \Delta \mathbf{m}_k,
    \boldsymbol{\lambda}_k,
    \mathcal{H}_k,
    \mathcal{A}
  \right),
  \quad
  \boldsymbol{\lambda}_{k+1}
  =
  \pi_{\mathrm{LLM}}(\mathbf{s}_k),
\end{equation}
where $J$ is the number of feedback metrics, $\Delta \mathbf{m}_k$ denotes recent metric changes, $\mathcal{H}_k$ is the history of previous metric and weight states, and $\mathcal{A}$ is the set of adjustable loss terms. The model architecture, optimizer, data pipeline, and set of training losses are not modified. DLC only changes the loss-weight vector used in the next training interval.

\subsection{DLC Feedback Loop}
\label{sec:dlc_feedback_loop}

DLC wraps an existing training pipeline with a periodic metric-guided feedback loop. At the beginning of training, we select a small fixed feedback subset from the training data and keep it unchanged throughout training to measure metric changes under a consistent condition. Using the same subset at every intervention reduces variation caused by changing sample composition, so metric changes more directly reflect the effect of model updates. This subset is used only for training-time controller feedback, while final performance is reported on the standard benchmark test sets.

At each intervention step, DLC evaluates the current checkpoint on the fixed feedback subset and computes task-relevant quality metrics. These metrics may include fidelity-oriented scores, such as PSNR and SSIM~\cite{wang2004image}; reference-based perceptual metrics, such as LPIPS~\cite{zhang2018unreasonable} or DISTS~\cite{ding2020image}; and no-reference quality metrics, such as NIQE~\cite{mittal2012making}, BRISQUE~\cite{mittal2012no}, CLIP-IQA~\cite{wang2023exploring}, MUSIQ~\cite{ke2021musiq}, or MANIQA~\cite{yang2022maniqa}, depending on the task. The resulting metric values, recent changes, and current loss weights are summarized into a structured feedback state. This state is then provided to an LLM controller, which returns the loss weights for the next training interval.

DLC does not optimize the feedback metrics directly as differentiable losses. Instead, it uses them as non-differentiable feedback to select the next loss-weight configuration, while gradient-based optimization still relies on the original differentiable losses of each pipeline. This is important because many restoration metrics are non-differentiable, unstable as direct objectives, or not designed as training losses.

\subsection{Feedback State and Controller Output}
\label{sec:feedback_state}

The controller input is a structured state rather than raw images. For each intervention step, it contains metric values, metric directions, recent metric changes, current loss weights, and adjustable loss terms. The metric direction specifies whether each metric should be maximized or minimized; for example, higher PSNR, SSIM, CLIP-IQA, MUSIQ, and MANIQA indicate better quality, while lower LPIPS, DISTS, NIQE, BRISQUE, and FID indicate better quality. Rather than collapsing all metrics into a single scalar score, DLC preserves their values and directions so that the controller can interpret each recent change as improvement or degradation.

The feedback state also includes a short task instruction that describes the overall restoration goal, such as improving visual quality according to the feedback metrics while avoiding severe degradation of already strong metrics. The controller uses this information together with recent metric and weight history to determine how the loss weights should change. The output is a structured dictionary of numeric loss weights:
\begin{equation}
  \boldsymbol{\lambda}_{k+1}
  =
  \left\{
    \lambda_{k+1}^{(1)}, \lambda_{k+1}^{(2)}, \ldots, \lambda_{k+1}^{(N)}
  \right\}.
\end{equation}
Only these numeric weights are used by the training loop. The controller may
also produce a short textual rationale for logging and analysis, but this rationale
is not used as an optimization signal.

This design allows DLC to handle heterogeneous metric profiles without requiring a manually designed analytic rule for every task. Instead of assuming a universal update formula, DLC uses the controller to interpret the metric state in the context of the available loss terms and recent training behavior. The role of the LLM is therefore narrow and operational: it maps a structured metric summary to the next loss-weight configuration. In our implementation, we instantiate the controller with LLaMA-3 8B~\cite{grattafiori2024llama} and keep the prompt template, decoding parameters, and structured output format fixed within each experiment. This fixed interface improves reproducibility by ensuring that the optimization process depends on structured numeric outputs rather than free-form textual responses.

\subsection{Domain Instantiations}
\label{sec:domain_instantiations}

DLC is instantiated over the existing loss space of each restoration domain. The controller formulation remains the same, but the adjustable losses and feedback metrics are task-dependent. This reflects the fact that different restoration tasks use different training objectives and evaluation protocols. Table~\ref{tab:domain_instantiations} summarizes the controlled losses for low-light enhancement, deraining, and real-world super-resolution. In all cases, DLC reweights only the existing loss terms of each original pipeline and does not introduce task-specific architectural changes. The controlled losses are inherited from the corresponding baseline objectives~\cite{zhang2025cwnet,dong2025channel,sun2025pixel} rather than newly designed for DLC. Thus, the comparison isolates the effect of the loss-weight decision policy.

\begin{table}[H]
  \centering
  \caption{Domain-specific DLC instantiations.}
  \label{tab:domain_instantiations}
  \small
  \renewcommand{\arraystretch}{1.08}
  \begin{tabular*}{0.92\linewidth}{@{\extracolsep{\fill}}lp{0.48\linewidth}@{}}
    \toprule
    Domain & Controlled losses \\
    \midrule
    Low-light Enhancement &
    pixel, perceptual, SSIM \\
    Deraining &
    reconstruction, consistency, perceptual, adversarial \\
    Real-world SR &
    L2, LPIPS, CSD \\
    \bottomrule
  \end{tabular*}
\end{table}

\subsection{Training Procedure}
\label{sec:training_procedure}

The overall DLC training procedure is summarized in Algorithm~\ref{alg:dlc_training}. The procedure only changes the loss weights at discrete intervention steps, while keeping the original model, optimizer, data pipeline, and training budget unchanged. Only the parsed numeric weights are passed to the training loop, while any textual rationale is used only for logging and analysis.

\vspace{-1.0em}
\begin{algorithm}[H]
  \caption{DLC Training Procedure}
  \label{alg:dlc_training}
  \small
  \begin{algorithmic}[1]
    \Require Model $f_{\theta}$, training data $\mathcal{D}_{train}$, feedback subset $\mathcal{D}_{fb}$, initial weights $\boldsymbol{\lambda}_0$, interval $T$
    \Ensure Trained restoration model $f_{\theta}$

    \State Initialize intervention index $k \leftarrow 0$
    \While{training budget is not exhausted}
    \State Train $f_{\theta}$ for one interval using $\boldsymbol{\lambda}_k$
    \State Evaluate the current checkpoint on $\mathcal{D}_{fb}$
    \State Compute feedback metrics $\mathbf{m}_k$
    \State Build feedback state $\mathbf{s}_k=(\mathbf{m}_k,\Delta\mathbf{m}_k,\boldsymbol{\lambda}_k,\mathcal{H}_k)$
    \State Query the LLM controller with $\mathbf{s}_k$
    \State Parse the output into $\boldsymbol{\lambda}_{k+1}$
    \State Set $k \leftarrow k+1$
    \EndWhile
  \end{algorithmic}
\end{algorithm}

\section{Experiments}
\label{sec:experiments}

\subsection{Experimental Setup}
\label{sec:experimental_setup}

We evaluate DLC on three restoration domains: low-light image enhancement~\cite{zhang2025cwnet}, deraining~\cite{dong2025channel}, and real-world super-resolution~\cite{sun2025pixel}. For each domain, the baseline is the corresponding original pipeline trained with fixed loss weights. DLC uses the same backbone, training data, optimizer, and set of training losses as the baseline, and changes only the loss-weight decision policy during training. For DLC, we use a fixed feedback subset of 10 training samples in each experiment. The same subset is evaluated at every intervention step to provide consistent metric feedback to the controller. All reported results are computed on the standard test sets.

For low-light image enhancement and deraining, we report PSNR, SSIM, LPIPS, NIQE, CLIP-IQA, and BRISQUE. For real-world super-resolution, we follow the broader evaluation protocol used in this domain and report PSNR, SSIM, LPIPS, DISTS, CLIP-IQA, NIQE, MUSIQ, MANIQA, and FID. Higher values are better for PSNR, SSIM, CLIP-IQA, MUSIQ, and MANIQA, while lower values are better for LPIPS, DISTS, NIQE, BRISQUE, and FID. We bold a DLC result only when it improves over the baseline according to the corresponding metric direction.

\subsection{Main Quantitative Results}
\label{sec:main_results}

\noindent\textbf{Low-light image enhancement.}
Table~\ref{tab:llie_results} reports results on LOL-v1~\cite{wei2018deep}, LOL-v2-Syn, and LOL-v2-Real~\cite{yang2020fidelity}. DLC improves most perceptual and no-reference quality metrics on LOL-v1, with only a minor PSNR drop. On LOL-v2-Syn, DLC provides clear gains in reference fidelity and perceptual similarity, improving PSNR, SSIM, LPIPS, and NIQE. The strongest improvement is observed on LOL-v2-Real, where DLC substantially increases PSNR from 27.39 to 30.74 while also improving structural and perceptual metrics. These results indicate that metric-guided loss reweighting can find a better fidelity--perceptual operating point than the baseline.

\begin{table}[H]
  \centering
  \caption{Low-light image enhancement results. DLC values are bolded when they improve over the baseline.}
  \label{tab:llie_results}
  \footnotesize
  \setlength{\tabcolsep}{3pt}
  \renewcommand{\arraystretch}{1.08}
  \begin{tabular*}{\linewidth}{@{\extracolsep{\fill}}llcccccc@{}}
    \toprule
    Dataset & Method & PSNR$\uparrow$ & SSIM$\uparrow$ & LPIPS$\downarrow$ & NIQE$\downarrow$ & CLIP-IQA$\uparrow$ & BRISQUE$\downarrow$ \\
    \midrule
    LOL-v1 & Base & 23.60 & 0.8482 & 0.1231 & 4.5780 & 0.3920 & 23.0670 \\
    LOL-v1 & DLC  & 23.50 & \textbf{0.8519} & \textbf{0.0974} & \textbf{4.4457} & \textbf{0.4630} & \textbf{19.9560} \\
    \midrule
    LOL-v2-Syn & Base & 25.50 & 0.9362 & 0.1195 & 4.2735 & 0.5261 & 13.3023 \\
    LOL-v2-Syn & DLC  & \textbf{26.07} & \textbf{0.9450} & \textbf{0.0395} & \textbf{4.1369} & 0.5025 & 13.3490 \\
    \midrule
    LOL-v2-Real & Base & 27.39 & 0.9007 & 0.0891 & 4.5415 & 0.4128 & 24.7530 \\
    LOL-v2-Real & DLC  & \textbf{30.74} & \textbf{0.9105} & \textbf{0.0622} & 4.5441 & \textbf{0.4711} & \textbf{24.2293} \\
    \bottomrule
  \end{tabular*}
\end{table}

\noindent\textbf{Deraining.}
Table~\ref{tab:deraining_results} summarizes the deraining results on Rain100L~\cite{yang2017deep} and RealRain1K-H~\cite{li2022toward}. On Rain100L, DLC improves PSNR, LPIPS, and BRISQUE, showing better reconstruction accuracy and perceptual quality. On RealRain1K-H, DLC improves SSIM, LPIPS, and CLIP-IQA, although PSNR and some no-reference metrics decrease. This suggests that DLC does not simply optimize a single metric, but shifts the model toward different operating points depending on the dataset and feedback signal.

\begin{table}[H]
  \centering
  \caption{Deraining results. DLC values are bolded when they improve over the baseline.}
  \label{tab:deraining_results}
  \footnotesize
  \setlength{\tabcolsep}{3pt}
  \renewcommand{\arraystretch}{1.08}
  \begin{tabular*}{\linewidth}{@{\extracolsep{\fill}}llcccccc@{}}
    \toprule
    Dataset & Method & PSNR$\uparrow$ & SSIM$\uparrow$ & LPIPS$\downarrow$ & NIQE$\downarrow$ & CLIP-IQA$\uparrow$ & BRISQUE$\downarrow$ \\
    \midrule
    Rain100L & Base & 33.2795 & 0.9537 & 0.0373 & 3.3761 & 0.7588 & 8.1748 \\
    Rain100L & DLC  & \textbf{33.5887} & 0.9526 & \textbf{0.0345} & 3.4173 & 0.7396 & \textbf{6.6552} \\
    \midrule
    RealRain1K-H & Base & 29.6702 & 0.9280 & 0.1720 & 10.4063 & 0.2349 & 51.2430 \\
    RealRain1K-H & DLC  & 29.1054 & \textbf{0.9318} & \textbf{0.1513} & 10.6044 & \textbf{0.2746} & 51.6570 \\
    \bottomrule
  \end{tabular*}
\end{table}

\noindent\textbf{Real-world super-resolution.}
Table~\ref{tab:sr_results} reports real-world super-resolution results on DRealSR~\cite{wei2020component} and RealSR~\cite{cai2019toward}. Across both datasets, DLC consistently improves PSNR, SSIM, LPIPS, and DISTS, indicating better reference fidelity and reference-based perceptual alignment under the same super-resolution backbone. DLC also improves MANIQA on both datasets and FID on DRealSR, while CLIP-IQA, NIQE, and MUSIQ show trade-offs, suggesting that DLC shifts the reconstruction--perception balance rather than uniformly improving all metrics. These results demonstrate that DLC can improve key fidelity and perceptual metrics without modifying the super-resolution architecture.

\begin{table}[H]
  \centering
  \caption{Real-world super-resolution results. DLC values are bolded when they improve over the baseline.}
  \label{tab:sr_results}
  \scriptsize
  \setlength{\tabcolsep}{2pt}
  \renewcommand{\arraystretch}{1.05}
  \resizebox{\linewidth}{!}{
    \begin{tabular}{llccccccccc}
      \toprule
      Dataset & Method & PSNR$\uparrow$ & SSIM$\uparrow$ & LPIPS$\downarrow$ & DISTS$\downarrow$ & CLIP-IQA$\uparrow$ & NIQE$\downarrow$ & MUSIQ$\uparrow$ & MANIQA$\uparrow$ & FID$\downarrow$ \\
      \midrule
      DRealSR & Base & 28.3197 & 0.7804 & 0.2960 & 0.2169 & 0.6970 & 6.1792 & 66.1088 & 0.6161 & 130.4879 \\
      DRealSR & DLC  & \textbf{29.1599} & \textbf{0.7974} & \textbf{0.2806} & \textbf{0.2124} & 0.6847 & 6.3306 & 64.7712 & \textbf{0.6166} & \textbf{130.1198} \\
      \midrule
      RealSR & Base & 25.5031 & 0.7418 & 0.2672 & 0.2044 & 0.6699 & 5.5009 & 70.1478 & 0.6551 & 124.1799 \\
      RealSR & DLC  & \textbf{26.1454} & \textbf{0.7490} & \textbf{0.2582} & \textbf{0.2003} & 0.6652 & 5.8489 & 68.4462 & \textbf{0.6627} & 124.5460 \\
      \bottomrule
    \end{tabular}
  }
\end{table}

\subsection{Qualitative Results}
\label{sec:qualitative_results}

\begin{figure}[H]
  \centering
  \includegraphics[
    width=\linewidth,
    trim=0cm 18.5cm 0cm 0cm,
    clip
  ]{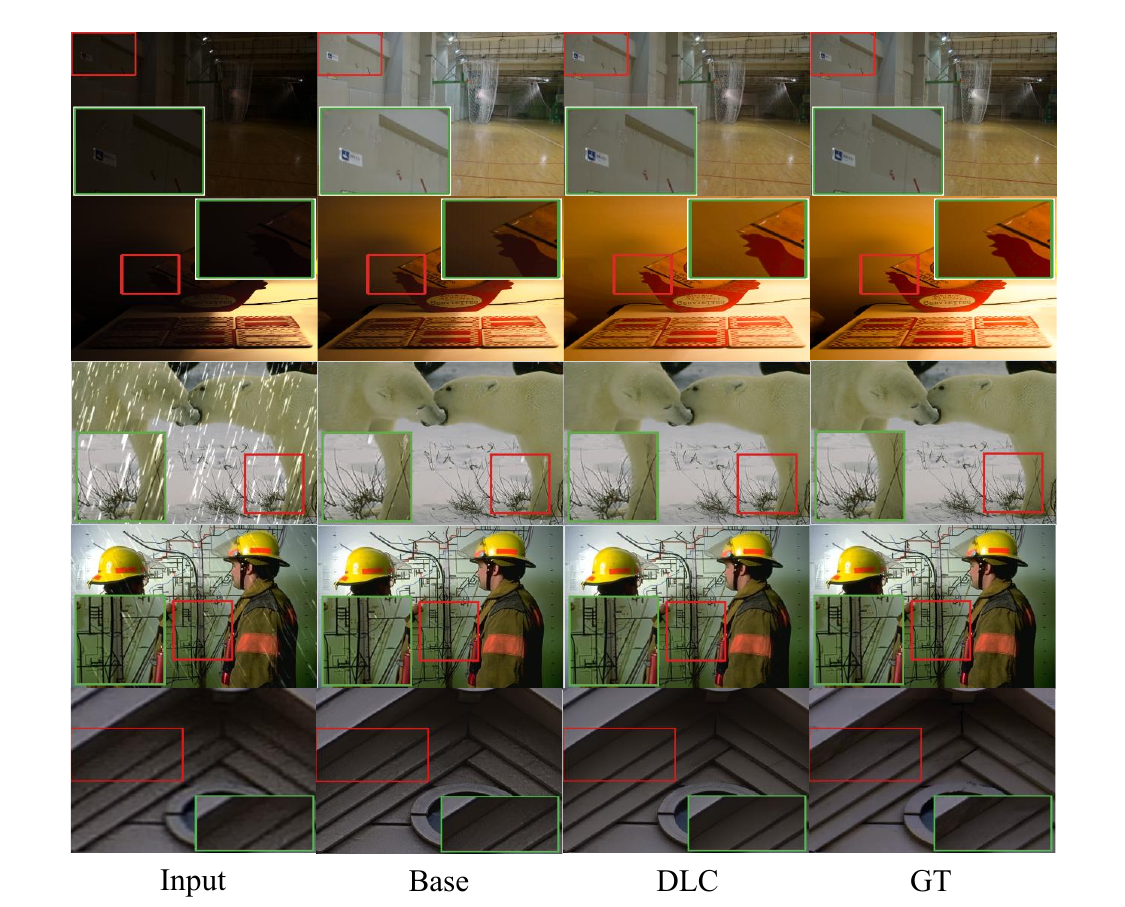}
  \caption{
    Qualitative comparison between the fixed-loss baseline and DLC across restoration domains. Zoomed crops highlight texture, structure, and perceptual differences.
  }
  \label{fig:qualitative_results}
\end{figure}

Fig.~\ref{fig:qualitative_results} compares the fixed-loss baseline and DLC across restoration domains using the same backbones. The zoomed crops highlight local differences in texture, structure, and perceptual quality.

\vspace{-1em}

The qualitative results provide visual examples of the metric-level trends observed in the quantitative evaluation. Compared with the baseline, DLC often yields clearer local structures or more favorable perceptual details in the highlighted crops. The improvements are not uniform across all regions, but they illustrate how metric-guided loss reweighting can shift the visual operating point without changing the underlying restoration architecture.

\subsection{Comparison of Loss-Weighting Policies}
\label{sec:ablation_policy}

To isolate the effect of the loss-weight policy, we compare DLC on LOL-v1 with
fixed-weight and rule-based baselines as well as GradNorm~\cite{chen2018gradnorm}, Uncertainty Weighting~\cite{kendall2018multi},
and CoV Weighting~\cite{groenendijk2021multi}. The fixed-weight variant uses DLC's final loss-weight vector
from the start of an independent training run, while the rule-based variant uses
a hand-designed metric-trend heuristic. All methods otherwise use the same
backbone, dataset, loss components, optimizer, and training budget.

Table~\ref{tab:ablation_policy} shows that different weighting policies lead to
different fidelity--perception trade-offs. DLC achieves the highest PSNR while
remaining competitive on SSIM and perceptual quality, whereas other methods
perform better on selected perceptual or no-reference metrics. These results
indicate that the resulting operating point depends on the feedback signal and
weight-update policy rather than on adaptive weighting alone. Unlike the other
adaptive methods, DLC determines weight updates from heterogeneous quality-metric
feedback together with recent loss-weight history.

\begin{figure}[H]
  \centering
  \includegraphics[
    width=\linewidth,
    trim=1.5cm 5.3cm 1.5cm 0cm,
    clip
  ]{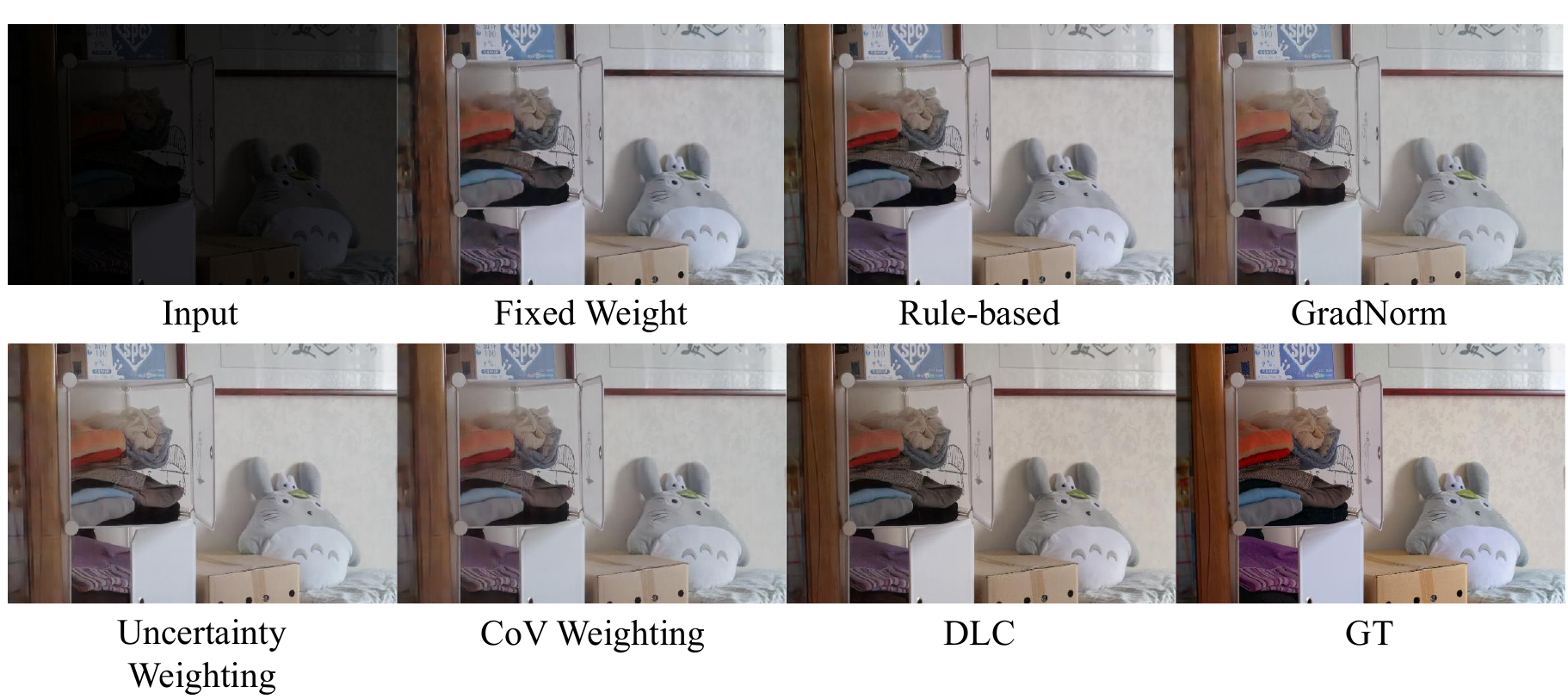}
  \caption{
    Qualitative comparison of loss-weighting policies on LOL-v1.
    The results illustrate the different visual characteristics induced by
    fixed, rule-based, and adaptive weighting strategies.
  }
  \label{fig:ablation_qualitative}
\end{figure}

Fig.~\ref{fig:ablation_qualitative} qualitatively compares all loss-weighting
policies and illustrates the visual differences induced by each strategy.

\begin{table}[t]
  \centering
  \caption{
    Comparison of loss-weighting policies on LOL-v1.
    All methods use the same backbone, dataset, loss components, optimizer,
    and training budget, and differ only in the loss-weight decision policy.
    The best result for each metric is bolded.
  }
  \label{tab:ablation_policy}
  \footnotesize
  \setlength{\tabcolsep}{3pt}
  \renewcommand{\arraystretch}{1.08}

  \begin{tabular*}{\linewidth}
    {@{\extracolsep{\fill}}lcccccc@{}}
    \toprule
    Method &
    PSNR$\uparrow$ &
    SSIM$\uparrow$ &
    LPIPS$\downarrow$ &
    NIQE$\downarrow$ &
    CLIP-IQA$\uparrow$ &
    BRISQUE$\downarrow$ \\
    \midrule

    Fixed weight
    & 23.11
    & \textbf{0.8548}
    & \textbf{0.0926}
    & 4.4830
    & 0.4580
    & 21.0060 \\

    Rule-based
    & 23.28
    & 0.8485
    & 0.0996
    & 4.4940
    & 0.4540
    & 20.8500 \\

    GradNorm
    & 22.81
    & 0.8473
    & 0.0934
    & \textbf{4.3496}
    & 0.4720
    & 19.6089 \\

    Uncertainty Weighting
    & 23.02
    & 0.8480
    & 0.0946
    & 4.3853
    & 0.4548
    & 19.9532 \\

    CoV Weighting
    & 22.77
    & 0.8482
    & 0.0938
    & 4.3791
    & \textbf{0.4845}
    & \textbf{19.2890} \\

    DLC (Ours)
    & \textbf{23.50}
    & 0.8519
    & 0.0974
    & 4.4457
    & 0.4625
    & 19.9560 \\

    \bottomrule
  \end{tabular*}
\end{table}

\vspace{-2em}

\subsection{Controller LLM Sensitivity}
\label{sec:controller_sensitivity}

We evaluate Qwen2.5-7B~\cite{qwen2025qwen25technicalreport},
Mistral-7B~\cite{jiang2023mistral7b},
and LLaMA-3 8B~\cite{grattafiori2024llama} under the same DLC setting,
and observe comparable performance across all three controllers, indicating that
DLC is not tied to a specific LLM backbone.

\begin{table}[t]
  \centering
  \caption{Controller LLM sensitivity on LOL-v1. Best values are bolded.}
  \label{tab:controller_sensitivity}
  \scriptsize
  \setlength{\tabcolsep}{2.5pt}
  \renewcommand{\arraystretch}{1.0}

  \begin{tabular*}{\linewidth}
    {@{\extracolsep{\fill}}lcccccc@{}}
    \toprule
    Controller &
    PSNR$\uparrow$ &
    SSIM$\uparrow$ &
    LPIPS$\downarrow$ &
    NIQE$\downarrow$ &
    CLIP-IQA$\uparrow$ &
    BRISQUE$\downarrow$ \\
    \midrule

    Qwen2.5-7B
    & 23.2304
    & 0.8505
    & 0.0978
    & \textbf{4.3938}
    & 0.4600
    & \textbf{19.4955} \\

    Mistral-7B
    & 23.2334
    & 0.8505
    & 0.0979
    & 4.4092
    & 0.4599
    & 19.5009 \\

    LLaMA-3 8B
    & \textbf{23.5045}
    & \textbf{0.8519}
    & \textbf{0.0974}
    & 4.4457
    & \textbf{0.4625}
    & 19.9560 \\

    \bottomrule
  \end{tabular*}
\end{table}

\subsection{Loss-Weight Trajectory Analysis}
\label{sec:weight_trajectory_analysis}

\begin{figure}[H]
  \centering
  \includegraphics[width=\linewidth,trim=0cm 1cm 0cm 1cm,clip]{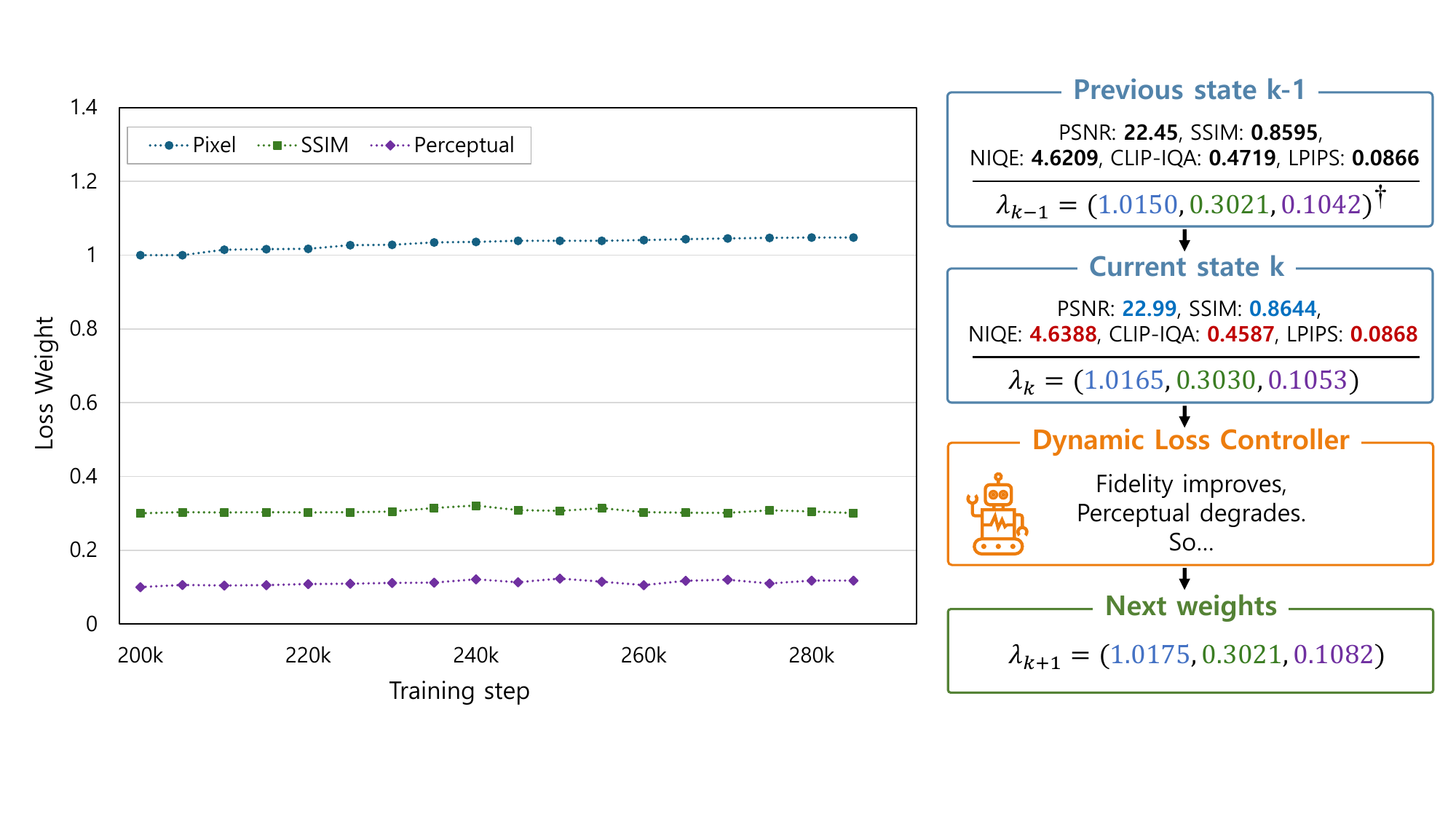}
  \caption{
    Loss-weight trajectory and controller decision example on LOL-v1.
    Left: representative trajectories of the pixel, SSIM, and perceptual loss weights over intervention steps.
    Right: an example controller decision at a single intervention step, where fidelity-oriented metrics improve while perceptual or no-reference feedback degrades.
    DLC adjusts the next loss weights according to the observed metric trade-off.
    $^\dagger$The loss-weight vector is ordered as $(\lambda_{\mathrm{pix}}, \lambda_{\mathrm{ssim}}, \lambda_{\mathrm{perc}})$.
  }
  \label{fig:weight_trajectory}
\end{figure}

To analyze how DLC changes the training objective over time, Fig.~\ref{fig:weight_trajectory}
shows the loss-weight trajectory and a representative controller decision on LOL-v1.
The pixel and perceptual weights gradually increase, while the structural weight remains
relatively stable, indicating that DLC mainly adjusts the reconstruction--perceptual balance.
When fidelity metrics improve but perceptual feedback degrades, the controller increases
the perceptual weight for the next interval. These results show that DLC dynamically
reweights existing losses according to observed metric trade-offs rather than following
a fixed update heuristic.
\vspace{-2em}

\subsection{Training-Time and Test-Time Cost}
\label{sec:cost_analysis}

DLC introduces additional cost only during training, due to periodic metric evaluation on the fixed feedback subset and controller inference. Table~\ref{tab:overhead_analysis} summarizes the measured overhead from representative runs across the three restoration domains. In low-light enhancement, each intervention takes approximately 7--8 seconds, resulting in a total overhead of 2.12--6.36 minutes. In deraining, each intervention takes approximately 7--9 seconds, with a total overhead of 1.49--2.17 minutes. Since the controller is invoked only at discrete intervals, the relative overhead remains below 0.2\% in these two domains.

For real-world super-resolution, DLC introduces a per-call overhead of 7.07 seconds and a total overhead of 0.68 minutes, corresponding to 1.69\% of the measured training time. Although the relative overhead is higher than in the other two domains, it is still incurred only during training. Importantly, DLC does not add any module to the restoration architecture and does not change the inference pipeline. Therefore, the final trained model has the same number of parameters and the same test-time cost as the baseline.

\begin{table}[H]
  \centering
  \caption{Training-time overhead of DLC measured from representative runs. The reported overhead is incurred only during training, and the final model has no additional test-time cost.}
  \label{tab:overhead_analysis}
  \footnotesize
  \setlength{\tabcolsep}{4pt}
  \renewcommand{\arraystretch}{1.08}
  \begin{tabular*}{\linewidth}{@{\extracolsep{\fill}}lccc@{}}
    \toprule
    Domain & Per-call overhead & Total overhead & Relative overhead \\
    \midrule
    Low-light enhancement & 7.49--7.71 s & 2.12--6.36 min & $< 0.2\%$ \\
    Deraining & 6.94--9.29 s & 1.49--2.17 min & 0.16--0.18\% \\
    Real-world SR & 7.07 s & 0.68 min & 1.69\% \\
    \bottomrule
  \end{tabular*}
  \vspace{-1em}
\end{table}

\section{Conclusion}
\label{sec:conclusion}

We presented DLC, a metric-guided dynamic loss controller for multi-objective restoration training. DLC preserves the existing model, optimizer, data pipeline, and training losses, and updates only the loss-weight vector at discrete training intervals using metric feedback from a fixed feedback subset. Through an LLM-based controller, DLC maps heterogeneous quality metrics, recent metric changes, and loss-weight history to adaptive loss reweighting.

Across low-light image enhancement, deraining, and real-world super-resolution, DLC improves key fidelity or perceptual metrics while making the remaining trade-offs explicit. The ablation on loss-weight update policies further indicates that the benefit comes not merely from changing weights, but from using metric feedback to guide the update direction.

DLC is a training-time plug-in, and the final model incurs no additional test-time cost. These results suggest that metric-guided loss control is a practical way to steer existing restoration pipelines toward more favorable operating points. Future work includes more systematic analysis of feedback subset selection, intervention frequency, and controller design.

\medskip
\footnotesize{\noindent \textbf{Acknowledgements.} This research was supported by the Republic of Korea Government, funded by the Ministry of Science and ICT (National Research Foundation of Korea (NRF) and the Institute for Information \& Communications Technology Planning \& Evaluation (IITP) [RS-2023-00242528(10\%), RS-2024-00437756(50\%)]), the Ministry of Education [BK21 FOUR:2120241015413(20\%), Glocal 30 Program(20\%)].}

%
%
\bibliographystyle{splncs04}
\bibliography{main}
\end{document}